# Who will stay? Using Deep Learning to predict engagement of citizen scientists

Alexander Semenov[1,3], Yixin Zhang[2] and Marisa Ponti[,2]

[1] Herbert Wertheim College of Engineering, University of Florida, FL, USA
[2] Department Two, Institution Two, City Two, Country Two
[3] Center for Econometrics and Business Analytics, Saint Petersburg University, Russia

E-mail: asemenov@ufl.edu

Citizen science and machine learning should be considered for monitoring the coastal and ocean environment due to the scale of threats posed by climate change and the limited resources to fill knowledge gaps. The preparation of high-quality, large volumes of training data, however, requires the participation of many volunteers. Understanding, sustaining, and improving volunteer engagement remains a long-standing challenge in citizen science. Analysis of citizen scientist engagement patterns may provide important insights into the organisation of citizen science projects. Using data from the annotation activity of citizen scientists in a Swedish marine project, we constructed Deep Neural Network models to predict forthcoming engagement. We tested the models to identify patterns in annotation engagement. Based on the results, it is possible to predict whether an annotator will remain active in future sessions. Depending on the goals of individual citizen science projects, it may also be necessary to identify either those volunteers who will leave or those who will continue annotating. This can be predicted by varying the threshold for the prediction.

The engagement metrics used to construct the models are based on time and activity and can be used to infer latent characteristics of volunteers and predict their task interest based on their activity patterns. They can estimate if volunteers can accomplish given number of tasks in a certain amount of time, identify early on who is likely to become a top contributor or identify who is likely to quit and provide them with targeted interventions. The novelty of our predictive models lies in the use of Deep Neural Networks and the sequence of volunteer annotations. A limitation of our models is that they do not use embeddings constructed from user profiles as input data, as many recommender systems do. We expect that including user profiles would improve prediction performance.

## Abstract

## 1. Introduction and background

Citizen science enables long-term and large-scale data collection and processing projects essential for tracking ecosystem health in ways that are unattainable through traditional ecological monitoring where all the work is done by a small team of scientists (Pecl *et al* 2019, Theobald *et al* 2015). In such large-scale projects, time and efforts spent by volunteering citizen scientists are crucial.

Marine citizen science (MCS), where volunteers at varying levels of expertise collaborate with professional scientists to study the marine environment, has been identified as one main component to gather empirical data and analyse data to generate knowledge about the coastal and ocean environment (Mission Board on Healthy Oceans, Seas, Coastal and Inland Waters 2020). MCS programs have grown in scope and number in recent years (Thiel *et al* 2014, Roy *et al* 2012), but they have lagged behind terrestrial citizen science and are underrepresented in the citizen science literature (Sandahl and Tøttrup 2020). Roy *et al* (2012) noted that, despite covering 70% of the earth's surface, marine environments were studied in only 14% of citizen science projects globally in 2012. There are several reasons for this result including the fact that the marine environment is inaccessible to many people and has had fewer naturalists and recorders than terrestrial flora and fauna (Garcia-Soto *et al* 2017). A recent study argued that MCS projects need to expand and diversify their volunteer base to be able to gather and process the data needed to realize the potential benefits of machine learning (ML) for marine ecological monitoring (Dalby *et al* 2021). Data-intensive MCS projects provide large datasets that can be used as training materials for ML. A subset of artificial intelligence, ML is achieved through adaptive algorithms that use large amounts of labelled data to autonomously detect patterns, make predictions, and recognize technical rules (Popenici and Kerr 2017 p. 2).

Given the scale of threats to marine environments and the relatively limited resources to fill knowledge gaps, citizen science and new technologies, such as ML, should be considered for ecological monitoring (Garcia-Soto *et al* 2021). To reap the benefits of ML, it is necessary to ensure the quality and volume of training data. Preparing high quality, large volume of training data requires significant volunteers' contributions. In order to develop reliable ML algorithms in citizen science projects, attracting and sustaining the engagement of a large number of citizen scientists is critical. Typically, citizen scientists are volunteering non-domain experts who use web applications to detect, annotate, or classify objects in images (Langenkämper *et al* 2019). Sustained engagement of citizen scientists leads to increased efficiency in terms of time and resources, and to quality outcomes that result from participant learning through repeated contributions (De Moor, Rijpma, Prats Lopez 2019). Nevertheless, retaining citizen scientists is challenging (De Moor *et al* 2019). According to previous studies, participation in citizen science projects plots along a solid core/periphery model, where the majority of participants contribute very little and a relatively small group contributes the most (Sauermann and Franzoni 2015). This has led to prior research exploring the motivations of volunteers (e.g. West *et al* 2021, Dalby *et al* 2021), how motivations change over time (e.g. Eveleigh *et al* 2014, Rotman *et al* 2014), or examining the quantity and quality of contributions over time (e.g. Aristeidou *et al* 2021, De Moor *et al* 2019).

By taking a step forward in this study, we propose ML models that predict volunteers' engagement patterns using the time spent and annotations made by volunteers in a MCS project in Sweden. Understanding engagement predictors can help inform recruitment and retention strategies and potential interventions (Rotman et al 2014). We were interested in answering the following research questions:

*How to predict volunteers' engagement patterns? How could their prior contributions predict their following contributions and what machine learning model would perform the best?*

Our goal is to provide citizen science project managers with models to predict engagement patterns among citizen scientists. The analysis of these patterns can provide important insights into the organization and management of citizen science project (Tupikina *et al* 2021). For example, we can identify potential top contributors earlier in the process, detect those who might stop contributing, and provide targeted interventions. Engagement predictions could also help explain the effects of different task design and interaction design. The ML models we propose are applicable to other individual citizen science projects that rely on persistent engagement.

## 1.1 Measuring engagement in citizen science

Engagement is a multidimensional concept that is understood, described, and measured in various ways across different research fields (Doherty and Doherty 2018). Quantitative research in the engagement literature has used interaction as a proxy for engagement, relating it to a scale of use, the unit of which is interaction (Doherty and Doherty 2018). Engagement involves time and attention contributed by volunteers to the tasks (Mao *et al* 2013). A main approach for measuring volunteer engagement quantitatively in online citizen science projects uses actual activity logs. Some researchers have used these logs to measure engagement in terms of duration of engagement, defined as the time spent by volunteers in a project (e.g. Aristeidou *et al* 2021, Ponciano and Brasileiro 2014), in terms of activity, defined as the number of submitted tasks (Cox *et al* 2015) and as the distribution of contributions among volunteers throughout a project (Sauermann and Franzoni 2015). In this study, we describe volunteers' engagement through their time spent on annotating and the number of annotations in a MCS project.

### 1.1.1 Predicting volunteers' engagement

While engagement metrics related to time and activity cannot explain why volunteers engage with a citizen science project, they can be used to infer latent characteristics of volunteers and predict their task interest based on their activity patterns (Ponciano and Brasileiro 2014). Predicting volunteers' engagement patterns can help project managers better organize engagement activities in their projects. For example, they can estimate how many tasks volunteers can accomplish in a certain amount of time, identify who are likely to become top contributors at an early stage and provide tailored training for them or identify who will quit and provide targeted interventions instead of carrying out general interventions for all volunteers. Using engagement metrics, we can predict the time of volunteers' activity and their frequency of annotations. Each annotator in an annotation project has multiple latent behavioural characteristics, such as interest in the project and willingness to continue contributing, which may fluctuate over time. Some annotators may gradually lose interest and leave the project, while others may become more engaged. Previous studies conducted in the field of citizen science examined the influence of age and task difficulty on retention for observing precipitation (Sheppard *et al* 2017) and stages of contributions of participants in citizen science projects (Fischer *et al* 2021).

A relatively novel methodology to develop prediction models is Deep learning (DL). DL describes a family of ML algorithms used successfully in multiple domains, including the construction of recommender systems (Emmert-Streib *et al* 2020). Recommender systems based on DL algorithms can be accurate in inferring latent characteristics of e-commerce customers and give good recommendations about customers' subsequent purchases, movies and so on (Wang *et al* 2021). DL models can often infer latent characteristics such as customers preferences based on the interactions between customers and items (such as products or movies), without requiring any additional data. Besides traditional recommender systems that are based on user-item matrix factorization, a widespread approach is to leverage sequences of users' actions for making a recommendation. Often, such recommender systems are based on recurrent neural network architectures, designed for handling sequential data (Wang *et al* 2021), such as time series.

Main tools of the DL are the Deep Neural Networks (DNN), that refer to a class of ML methods that are constructed from layers of *neurons* connected to each other. DNN may be mathematically represented as a nonlinear function, mapping the input data to output variables. Its *learning* is to find parameters of this function that return accurate output values corresponding to the given input. Modern DNN architectures contain millions of parameters and require powerful computational resources for training.

## 2. Research Context: the Koster Seafloor Observatory for biodiversity research

The Koster Seafloor Observatory (herein KSO) was our research setting. KSO is an open-source approach to analyse large amounts of subsea movie data for marine ecological research. KSO is available on the Zooniverse, the Internet's largest citizen science project platform. The project aims to investigate marine biodiversity through analyzing deep-water recordings from the Kosterhavets National Park on the west coast of Sweden. The Kosterhavets National Park is Sweden's first marine national park, and scientists have been using remotely-operated vehicles (ROVs) to record deep-water videos in this marine protected area. Such videos help scientists to investigate the spatial-temporal distribution of relative abundance of habitat-building species, such as cold-water corals. Thanks to scientists' continuous efforts, there are more than 20 years of recordings from the seafloor in this area. However, the massive number of videos present great challenges for analysis (Anton *et al* 2021).

In the KSO project, scientists invited citizen scientists to identify the species of animals recorded in video clips and still images (Figure 1) and specify the number of individuals of the taxon selected and the time (in seconds) when any of the individuals fully appears on the screen. To ensure the quality of annotations, the same video clip is analysed by multiple citizen scientists, and only if a certain level of consistency is reached among the annotation results, the specific video clip is considered properly annotated. In certain cases, experts also check citizen annotations to ensure the quality of the annotations.

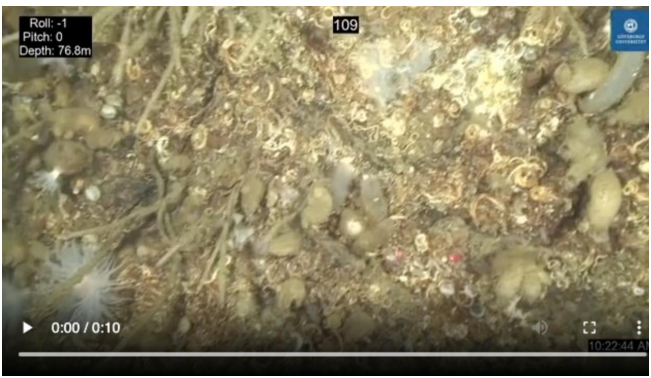

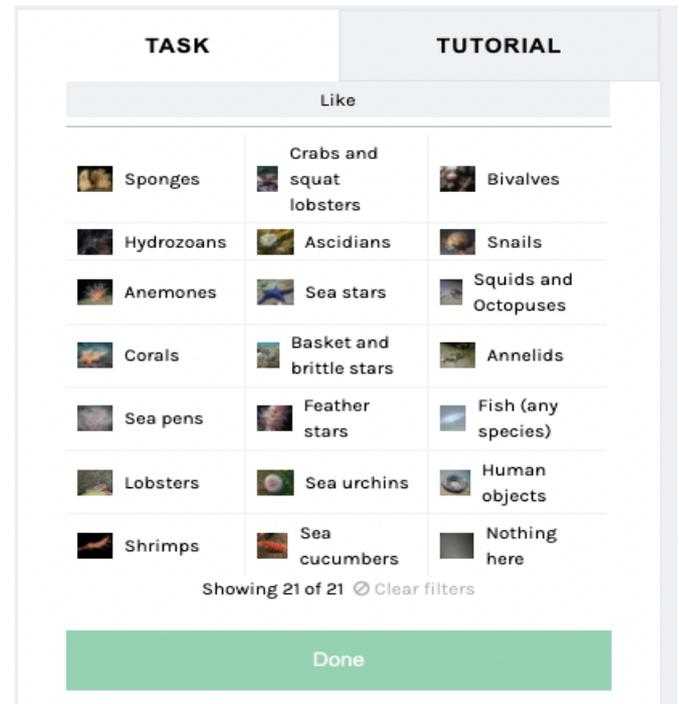

*Figure 1 Koster annotation interface*

The video clips with annotations added by citizen scientists are then used as training data for the development of ML models. KSO has resulted in an ML model with adequate performance, which has been entirely trained with annotations provided by citizen scientists. Therefore, Anton *et al (*2021) pointed out that the combination of open-source tools, citizen science, ML and high-performance computational resources is key to successfully analysing large amounts of underwater imagery in the future.

## 3. Model Development

### 3.1 Data collection and data description

We accessed a dataset of citizen scientists' annotation data in KSO by accessing Zooniverse data through Panoptes client. We used Python programming language to analyse the data. As we did not conduct research involving humans (such as experiments) or animals, and we did not process personal and sensitive data, this type of study did not require ethical approval under the Swedish law concerning the ethical review of research involving humans (2003:460).

The collected dataset includes 110953 annotations from 6488 participants over a period from October 1, 2019 to October 21, 2021. Small numbers of citizen scientists contribute most annotations, leading to skewed participation. The top 20 users contributed 20710 annotations, or 18.67% of the 110953 annotations. Among the 6488 users who contributed to the annotations, 4085 users logged into Zooniverse with user ids, 2403 users did not log into Zooniverse and were tracked through cookies. Zooniverse allows users to work on annotations without logging into the system. Users who logged into the system contributed on average 20.99 annotations, and users who did not log into the system contributed on average 10.53 annotations. Two-tailed t-test (p-value < 0.0001) indicates those who logged in contributed significantly more annotations than those who did not log in. The descriptive statistics suggests volunteers differ significantly in terms of engagement. In order to better plan the project and develop targeted intervention measures, it is important to predicate volunteer's engagement patterns.

*3.2 Modelling the problem of predicting volunteer engagement*

Participating users sequentially perform the annotations, which results in time series with annotations for each user. For each user $i$ we denote annotation sequence of $T$ annotations as $A^i = \{a_1^i, a_2^i, \ldots, a_t^i, \ldots, a_T^i\}$, where $a_t^i$ is the annotation made by user $u$ at time $t$. Further, we use $A_{s,e}^i$ to represent annotation subsequence starting at timestamp $s$ and ending at $e$. Characteristics of these sequences are different for each user; some users perform annotations regularly, while others perform few annotations per day. In the current study, we adopt an engagement metric, operationalizing the number of contributions that a user would make in the future. As in Mao et al 2013, we use a binary variable indicating whether a user would perform more than a specified number of annotations $\gamma$ during a period of time, i.e. $I_F[|A_{s,e}^i| > \gamma]$ as an engagement metric, where $I_F[.]$ is an indicator function

Consider a dataset $\mathcal{D} = \{(x_i, y_i)\}_{i=1}^n$, where $x_i \in \mathcal{X}$ is the $i$th observed variable, $y_i \in \mathcal{Y}$ is the corresponding $i$th label. Our goal is to learn a function $\hat{y}_j = f(x_j, \theta)$, where $\hat{y}_j$ is the predicted class, and $\theta$ is a vector of parameters. In order to find the vector of optimal parameters $\theta$, we need to minimize the loss function $\mathcal{L}(y, \hat{y})$ between the actual and predicted classes. There are many ways to model function $f$, in the current study we propose to utilize several DNN architectures for modelling this function. The process of finding optimal parameters given the data is referred to as *model training*. After training, the model should be evaluated by calculating its performance characteristics. In order to first train the model and then evaluate the model, the dataset is typically split into two parts: training dataset, and testing dataset (typically 80% and 20%); the model is trained on the training dataset, and evaluation is performed on testing dataset that the model has not "seen" during the training.

We define the prediction challenge as follows: given the current information on a volunteer's past contributions $A_{s_p,e_p}^i$ will s/he maintain a specific engagement level within a given number of tasks or minutes of time in the future? Then, inspired by Mao *et al* (2013), the engagement prediction problem can be formalized as a binary classification problem predicting $I_F[|A_{s_f,e_f}^i| > \gamma]$. Here, $s_p, e_p, s_f, s_f$ correspond to start and end timestamps in the past and in the future, and $s_p < e_p < s_f < s_f$. The training dataset is constructed from annotation histories in past $\{A_{s_{p,i},e_{p,i}}^i\}_{i=1}^N$, and the testing dataset is constructed from $\{A_{s_{f,i},e_{f,i}}^i\}_{i=1}^N$ where $N$ is the number of users. As noted by Mao *et al* (2013), most users have specific frequency patterns in their annotations. For example, most volunteers do the work only during the day, with breaks for a night or for a weekend. Further, sequential activities such as click logs or sentences in text often exhibit long-term dependencies, where only recent items play an important role in future activities. Drawing from the previous study, we split user activities into "sessions" of variable length, with longer breaks between the sessions. Each session itself is composed of a further sequential activities that often exhibit long-term dependencies, where only recent items play an important role on future activity. We split user annotation histories into sessions, where each session is a subset of a set $A_{s,e}^i$ such that there is at least a 30-minute time difference between the last annotation in the previous session, and the first annotation in the subsequent session.

*3.2.1 Feature engineering and model architecture*

As we do not have any user-related information such as user demographics, we utilize the annotation activity log to generate the feature vector $X$ for subsequent prediction. We generate the following features: 1) average number of annotations per session, 2) average number of annotations in five recent sessions, 3) average time spent between two subsequent annotations in a session for all past sessions, 4) average time spent between two subsequent annotations in a session for all past tasks in current session 5) total number of completed annotations, 6) number of completed annotations in the current session, and 7) flag, if the user was logged in. Besides these features, we append to the beginning of the feature vector time differences between M+1 most recent annotations made by a user in a session (see Figure 3) for capturing implicit patterns in annotation time difference, we experiment with M = {5, 10}, resulting in five and ten additional elements in the feature vector, respectively.

Elements of the dataset $\mathcal{D}$ are generated by moving a sliding window from the beginning of a session to its end; each new annotation leads to adding an item to the dataset. An example of the session and one training item is shown in Figure 2. The picture shows part of the activity log of one user; the picture contains two sessions (with 12h time difference); and a sliding window containing time differences between six annotations with four annotations that will still be made within the current session. The next item will have a sliding window shifted forward for one element, modified features with updated averages, and decremented value *y*. Resulting dataset had 24608 entries.

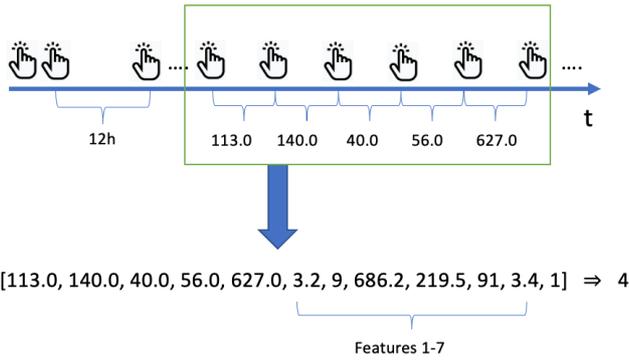

*Figure 2 Session and example of one input vector*

We propose two deep neural network architectures as the main prediction models. The first architecture of the neural network (referred to as LSTM-net) is presented in Figure 3. It utilizes recurrent neural network and is built based on Long-Short Term Memory (LSTM) cells. It has two inputs: feature vector and series of time steps are sent to the model independently; then, output of dense layer is concatenated with LSTM output, followed by two dense layers with rectified linear unit (ReLU) activation and sigmoid output. The second architecture (referred to as DNN-net) is a feedforward DNN; the first three layers are fully connected layers having ReLU activation function, and the last layer is a sigmoid neuron returning value between 0 and 1. The architecture is illustrated in Figure 4.

### 3.2.2 Model evaluation

We evaluate the model on nine gammas ($\gamma$).: {2, 5, 8, 10, 15, 20, 25, 50, 75}. As we have an imbalanced dataset, we use Area Under the Curve (AUC) characteristic to evaluate the model; Receiver Operating Characteristic (ROC) curves are insensitive to changes in class distribution, and AUC represents aggregated measure of classifier performance (Fawcett, 2006). Uninformed classifier results in AUC score 0.5, whereas perfect classifier results in AUC equal to 1.0. the AUC may be interpreted as a probability that the classification model will rank a randomly chosen positive instance higher than a randomly chosen negative instance (Fawcett, 2006). We use a Random Forest classifier and Logistic Regression as baselines.

*Table 1 AUC results, 4-fold forward chaining. For M = 5.*

| $\gamma$ | LSTM-net | DNN-net | RF | LR |
| --- | --- | --- | --- | --- |
| 2 | **0.679±0.036** | 0.665±0.027 | 0.538±0.017 | 0.514±0.010 |
| 5 | **0.675±0.029** | 0.651±0.019 | 0.596±0.014 | 0.544±0.016 |
| 8 | **0.672±0.037** | 0.660±0.024 | 0.628±0.022 | 0.595±0.015 |
| 10 | **0.669±0.016** | 0.668±0,.014 | 0.648±0.020 | 0.611±0.038 |
| 15 | 0.659±0.008 | **0.662±0.018** | 0.661±0.017 | 0.613±0.016 |
| 20 | **0.695±0.023** | 0.674±0.039 | 0.679±0.010 | 0.616±0.010 |
| 25 | 0.649±0.046 | **0.696±0.017** | 0.688±0.019 | 0.618±0.015 |
| 50 | 0.757±0.135 | **0.796±0.128** | 0.780±0.114 | 0.702±0.035 |
| 75 | 0.802±0.134 | **0.836±0.109** | 0.764±0.155 | 0.629±0.155 |

We use forward chaining for model evaluation: we split the dataset into five equal parts, and construct training and testing data using sliding window approach, as illustrated in Figure 5. We use past data for training and subsequent chunk of data for testing. Blue parts represent training data, while green parts represent testing data. We implemented neural networks using Keras with Tensorflow Backend. We train all DNNs using Adam optimization algorithm for ten epochs (these and other parameters were found via hyperparameter tuning). Tables 1 and 2 contain averaged results of AUC characteristic for 5 and 10 feature timesteps, respectively, the best result is highlighted in boldface font. We can see that proposed architectures consistently outperform the baselines (Random Forest classifier (referred to as RF) with 50 trees, and a Logistic Regression model (referred to as LR)).

*Table 2 AUC results, 4-fold forward chaining. For M = 10*

| $\gamma$ | LSTM-net | DNN-net | RF | LR |
| --- | --- | --- | --- | --- |
| 2 | **0.672±0.027** | 0.658±0.015 | 0.628±0.022 | 0.596±0.021 |
| 5 | 0.817±0.166 | **0.828±0.084** | 0.766±0.157 | 0.630±0.101 |
| 8 | 0.770±0.114 | **0.791±0.101** | 0.779±0.113 | 0.701±0.036 |
| 10 | **0.665±0.048** | 0.661±0.043 | 0.596±0.012 | 0.549±0.012 |
| 15 | **0.689±0.004** | 0.662±0.037 | 0.680±0.018 | 0.627±0.013 |
| 20 | 0.677±0.014 | **0.692±0.005** | 0.679±0.016 | 0.619±0.016 |
| 25 | **0.673±0.020** | 0.658±0.010 | 0.639±0.028 | 0.621±0.031 |
| 50 | 0.654±0.022 | **0.662±0.016** | 0.660±0.017 | 0.618±0.011 |
| 75 | **0.687±0.022** | 0.636±0.031 | 0.545±0.009 | 0.510±0.006 |

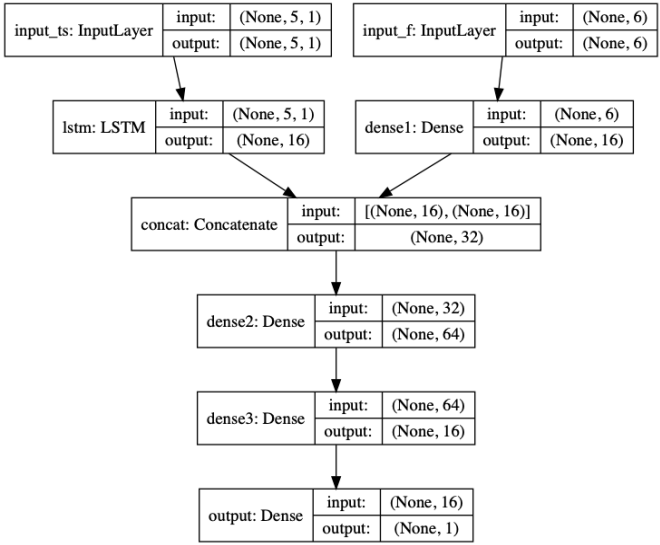

*Figure 3 Deep Neural Network architecture (1)*

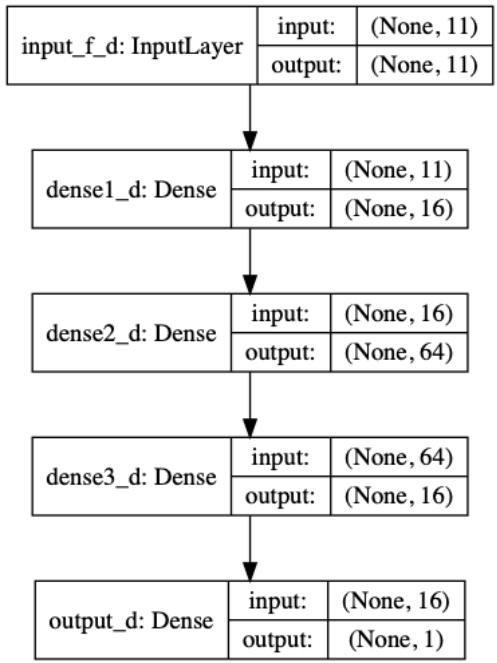

*Figure 4 Deep Neural Network architecture (2)*

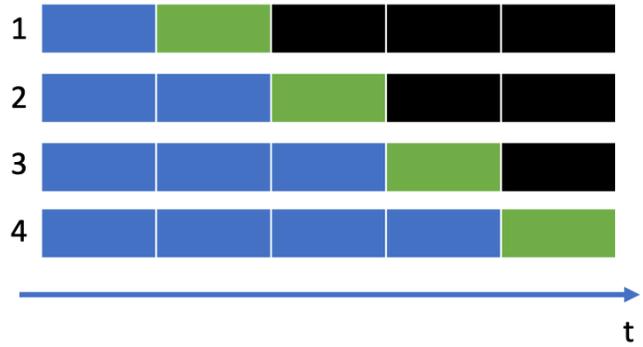

*Figure 3 Forward chaining for model evaluation (Blue parts represent training data, while green parts represent testing data)*

Tables 1 and 2 show that the proposed architectures always result in AUC score greater than 0.5 (that would correspond to a random guess), suggesting that the performance of the predictive model is better than an uninformed classifier. We can also see that standard deviation of the AUC is low, meaning that the model achieves good prediction results on all splits. We can also observe that the model with five timestep features results in, in general, better performance. Overall, the results indicate that the models are capable to predict whether an annotator will still be active in the current session. Depending on the requirements it is possible to perform the predictions for different values of γ. For example, it might be needed to find, if specific annotator would do more than six annotations in the current session; or whether s/he will do more than that.

Current prediction models are evaluated on the short-term predictions. Since the the AUC results averaging are stable, we expect that longer term engagement can also be predicted. Depending on the goals of the citizen science project managers, it may be important to identify among the volunteers who are going to leave, or who are going to continue the annotation; this can be done by varying the threshold of the prediction.

## 4. Discussion

We presented the development of predictive models of engagement, using data log capturing the activity of citizen scientists in the KSO, a MCS project. Participation in this project is skewed, and the fact that a small number of active participants contribute the highest number of annotations not only has a great influence on the resulting dataset (Cooper *et al* 2017) that feeds into the ML classifying algorithm, but also has consequences for the sustainability of the project in a longer term.

Besides being trained on the relatively small dataset of the KSO project, our models have also achieved desirable results when predicting near disengagement using Galaxy Zoo, a large-scale citizen science project (details of the prediction using Galaxy Zoo dataset can be provided on request). The development of these predictive models is significant because we see several practical applications in the KSO project. For example, identifying volunteers nearing disengagement, or exhibiting active behaviors, can inform interventions aimed at extending or retaining their engagement. We might be able to entice early dropouts back into participation through interactive educational modules that require them to identify species, spot differences in the biological composition, and then determine whether or not these differences are natural or caused by human activity. There can also be interventions targeting volunteers who gradually lose interests, such as achievement-based badges. In Mao *et al* (2013)'s view, if even a small fraction of those who leave respond to the interventions by staying and continuing to annotate, or by returning to the KSO project with a higher likelihood of involvement, the project can reap significant benefits from predicting engagement and disengagement.

The study has limitations. While we assert that these predictions can be useful, they are abstracted away from participant demographics, such as age, gender, and education level, which can be associated with retention. In particular, age can be a significant predictor of retention (Sheppard *et al* 2017). Such features were not available in the current dataset, so we could not use them when developing the models. If such participant features are included in future algorithm development, they will provide useful information for interventions that are aimed at encouraging certain age groups to participate more actively.

## 5. Conclusion

The proposed models can be integrated into hybrid evaluation systems that combine algorithmic prediction and qualitative data on motivation and participation experience to sustain engagement and reduce dropout rates. Volunteers are diverse and their past annotation behaviours should be cross-linked to their individual characteristics, such as capacities and motivations.

Previously, a statistical model has been proposed to predict how volunteers will engage in a citizen science project (Mao et al 2013). This study relied on a traditional development approach based on the construction of a large number of features. Our models differ from this example because we use DNN for prediction and consider the sequences of users' annotations. As previously mentioned, a limitation of our models is that they do not use embeddings constructed from user profiles as input data, as many recommender systems do. Future development of the predictive models could include user profile because descriptive statistics suggest that volunteers who log into the system contribute significantly more annotations than those who do not. We also expect that more advanced recurrent neural network models trained on large datasets would increase prediction performance.

Further studies should include task features and more thorough model evaluation, considering long-term prediction performance. In addition, it would be of interest to model the problem as a regression problem, not as a binary classification problem, in order to predict future contributions.

## Acknowledgements

We are indebted to the research team of the Koster Seafloor Observatory and the citizen scientists, without whom this study would not have been possible. This study has been funded by…